\documentclass[letterpaper, 10 pt, conference]{ieeeconf}
\IEEEoverridecommandlockouts    % This command is only needed if
\usepackage{graphics}           
\usepackage{times}              
\usepackage{amsmath}            
\usepackage{amssymb}            
\usepackage{graphicx}
\usepackage{algorithm}
\usepackage[noend]{algpseudocode}
\usepackage{booktabs}
\usepackage{color}
\usepackage{listings}
\usepackage{subfiles}
\usepackage{hyperref}
\usepackage[nocompress]{cite} % sorts citations automatically, e.g., "[1],[2],[3]" instead of "[2],[3],[1]".
\definecolor{instructioncolor}{rgb}{.5,.5,.5}

\usepackage[font=small]{caption}

\def\figref#1{Fig.~\ref{#1}}
\def\tabref#1{Tab.~\ref{#1}}
\def\eqref#1{Eq.~(\ref{#1})}

\makeatletter
\usepackage{xspace}
\DeclareRobustCommand\onedot{\futurelet\@let@token\@onedot}
\def\@onedot{\ifx\@let@token.\else.\null\fi\xspace}

\def\etal{{et al}\onedot}
\makeatother

\usepackage{array}
\newcolumntype{L}[1]{>{\raggedright\let\newline\\\arraybackslash\hspace{0pt}}m{#1}}
\newcolumntype{C}[1]{>{\centering\let\newline\\\arraybackslash\hspace{0pt}}m{#1}}
\newcolumntype{R}[1]{>{\raggedleft\let\newline\\\arraybackslash\hspace{0pt}}m{#1}}

\title{\LARGE \bf  Self-Supervised Moving Object Segmentation\\of Sparse and Noisy Radar Point Clouds}

\author{
Leon Schwarzer \qquad \qquad  Matthias Zeller \qquad \qquad  Daniel Casado Herraez \\[4pt]
Simon Dierl \qquad \qquad  Michael Heidingsfeld \qquad \qquad  Cyrill Stachniss% <-this % stops a space
  \thanks{L. Schwarzer is with CARIAD SE, Germany, and with TU Dortmund University, Germany. M. Zeller and D. Casado Herraez are with CARIAD SE, Germany, and with the University of Bonn, Germany. S. Dierl is with TU Dortmund University, Germany. M. Heidingsfeld is with CARIAD SE, Germany. C. Stachniss is with the University of Bonn, Germany, and with the Lamarr Institute for Machine Learning and Artificial Intelligence, Germany.}%
}

\usepackage{colortbl, xcolor}
\usepackage{tikz}

\definecolor{tablegrey}{RGB}{210,210,210}
\newcommand{\shaderow}{\rowcolor{tablegrey}}

\newcommand\copyrighttext{%
    \footnotesize 
    \color{gray}
    \textcopyright  
    2025
    \ IEEE. Personal use of this material is permitted.  Permission from IEEE must be obtained for all other uses, in any current or future media, including reprinting/republishing this material for advertising or promotional purposes, creating new collective works, for resale or redistribution to servers or lists, or reuse of any copyrighted component of this work in other works.
}

\begin{document}
\thispagestyle{empty}
\pagestyle{empty}
\maketitle

%%%%%%%%%%%%%%%%%%%%%%%%%%%%%%%%%%%%%%%%%%%%%%%%%%%%%%%%%%%%%%%%%%%%%%%%%%%%%%%%
\begin{abstract}

  Moving object segmentation is a crucial task for safe and reliable autonomous mobile systems like self-driving cars, improving the reliability and robustness of subsequent tasks like SLAM or path planning. While the segmentation of camera or LiDAR data is widely researched and achieves great results, it often introduces an increased latency by requiring the accumulation of temporal sequences to gain the necessary temporal context. Radar sensors overcome this problem with their ability to provide a direct measurement of a point's Doppler velocity, which can be exploited for single-scan moving object segmentation. However, radar point clouds are often sparse and noisy, making data annotation for use in supervised learning very tedious, time-consuming, and cost-intensive. To overcome this problem, we address the task of self-supervised moving object segmentation of sparse and noisy radar point clouds. We follow a two-step approach of contrastive self-supervised representation learning with subsequent supervised fine-tuning using limited amounts of annotated data. We propose a novel clustering-based contrastive loss function with cluster refinement based on dynamic points removal to pretrain the network to produce motion-aware representations of the radar data.
  Our method improves label efficiency after fine-tuning, effectively boosting state-of-the-art performance by self-supervised pretraining.

\end{abstract}

% copyright notice
\begin{tikzpicture}[remember picture,overlay]
        \node[anchor=south,yshift=10pt] at (current page.south) {\parbox{\dimexpr0.75\textwidth-\fboxsep-\fboxrule\relax}{\copyrighttext}};
\end{tikzpicture}

%%%%%%%%%%%%%%%%%%%%%%%%%%%%%%%%%%%%%%%%%%%%%%%%%%%%%%%%%%%%%%%%%%%%%%%%%%%%%%%%
\section{Introduction}
\label{sec:intro}

%%%%%%%%%%%%%%%%%%%
The ability of self-driving cars to reliably differentiate between moving and stationary objects is crucial for safe autonomous driving. The knowledge of which parts of the environment are actually moving can improve the reliability and robustness of subsequent tasks like SLAM~\cite{herraez2025ral} or path planning and facilitates navigation and collision avoidance in dynamic environments. Current autonomous systems are equipped with different sensors, including cameras, LiDAR, and radar sensors. To differentiate moving and stationary objects, the extensively researched cameras and LiDAR sensors utilize temporal sequences of data, which introduce an increased latency, further delaying the rest of the autonomous software stack. In contrast, radar sensors offer a per-point measurement of the Doppler velocity, incorporating instantaneous motion information that can be exploited to perform motion segmentation on the basis of a single scan~\cite{zeller2023icra,pang2024radarmoseve}. Radar sensors are also more robust against adverse weather conditions, but face the problem of sparse and noisy measurements. The high noise level makes the task of moving object segmentation more difficult, rendering most threshold-based approaches unacceptable \cite{scheiner2021object} and motivating machine learning methods. However, the sparsity of typical radar point clouds complicates the manual data annotation process necessary for supervised learning methods. This motivates the use of a self-supervised approach, which is able to solve complex tasks by utilizing large amounts of unannotated data. While there are multiple self-supervised learning approaches for radar~\cite{zhou2023itsc,akita2023itsc}, they do not consider moving object segmentation.

\begin{figure}
    \centering
    \includegraphics[width=\linewidth]{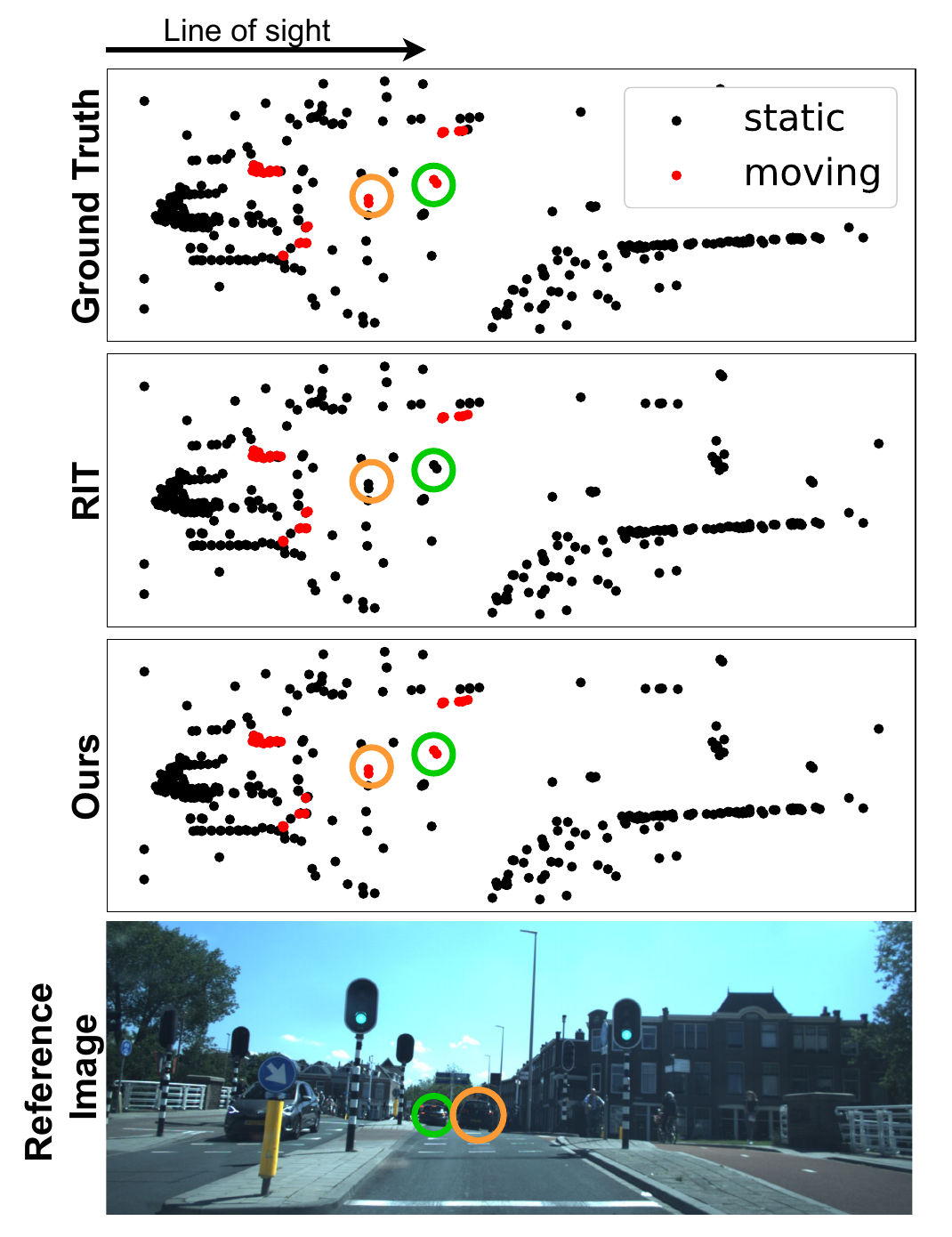}
    \caption{Our method improves the moving object segmentation performance of the underlying radar instance transformer architecture~\cite{zeller2024tor} by utilizing unannotated data to pretrain the network. The reference image shows that our approach correctly predicts the two cars in front as moving, in contrast to the base architecture trained from scratch. Best viewed in color.}
    \label{fig:motivation}
\end{figure}

%%%%%%%%%%%%%%%%%%%
In this paper, we specifically tackle the problem of moving object segmentation of sparse and noisy radar point clouds as illustrated in~\figref{fig:motivation}. This task requires differentiation between stationary and moving objects. In contrast to the task of moving instance segmentation, we do not group the moving points into instances but focus entirely on the semantic segmentation. Instead of using sequences of multiple radar scans, we exploit single radar point clouds, including the Doppler velocity and radar cross section. While most state-of-the-art methods for moving object segmentation rely on fully supervised training of neural networks, we employ self-supervised contrastive representation learning with supervised fine-tuning. The goal is to investigate whether the effect of using self-supervised pretraining to improve a neural network's performance observed in LiDAR-based scene understanding \cite{huang2021iccv,wang2024cvpr} can be transferred to the domain of radar-based sensing.
%%%%%%%%%%%%%%%%%%%
We analyze the effect of self-supervised pretraining by taking an existing fully supervised approach and using self-supervised contrastive representation learning to pretrain the backbone to produce motion-aware representations. Subsequently, we fine-tune the network in a supervised fashion. 

The main contribution of this paper is a novel contrastive loss function for self-supervised contrastive learning of motion-aware representations for automotive radar data. While most existing contrastive loss functions focus on semantic characteristics and were developed for LiDAR data, our problem requires a loss function focusing on motion information that is designed to cope with the specific characteristics of radar data. The proposed loss function utilizes unsupervised and algorithmic methods to group points in a radar scan into refined clusters whose distance in the representation space is taken as a similarity metric for the contrastive nature of the learning approach. We integrate this novel loss function into a framework for self-supervised contrastive pretraining to complete our novel self-supervised approach to the task of radar-based moving object segmentation.

%%%%%%%%%%%%%%%%%%%

In sum, we make three key claims: 
(i) Our approach is able to improve the label efficiency of the underlying network architecture when compared to training the model from scratch;
(ii) Using only 1\% of annotations from the View-of-Delft dataset, our approach outperforms the self-supervised baseline of RaFlow~\cite{ding2022ral} by over 25 absolute percentage points;
(iii) The clustering, as well as the cluster refinement using the dynamic points removal filtering, are essential to successfully group and differentiate the classes of moving and non-moving points in the representation space and thereby improve the performance.
All of the three claims are backed up by our experimental evaluations.

\begin{figure*}[t]
  \centering
  \includegraphics[width=\linewidth]{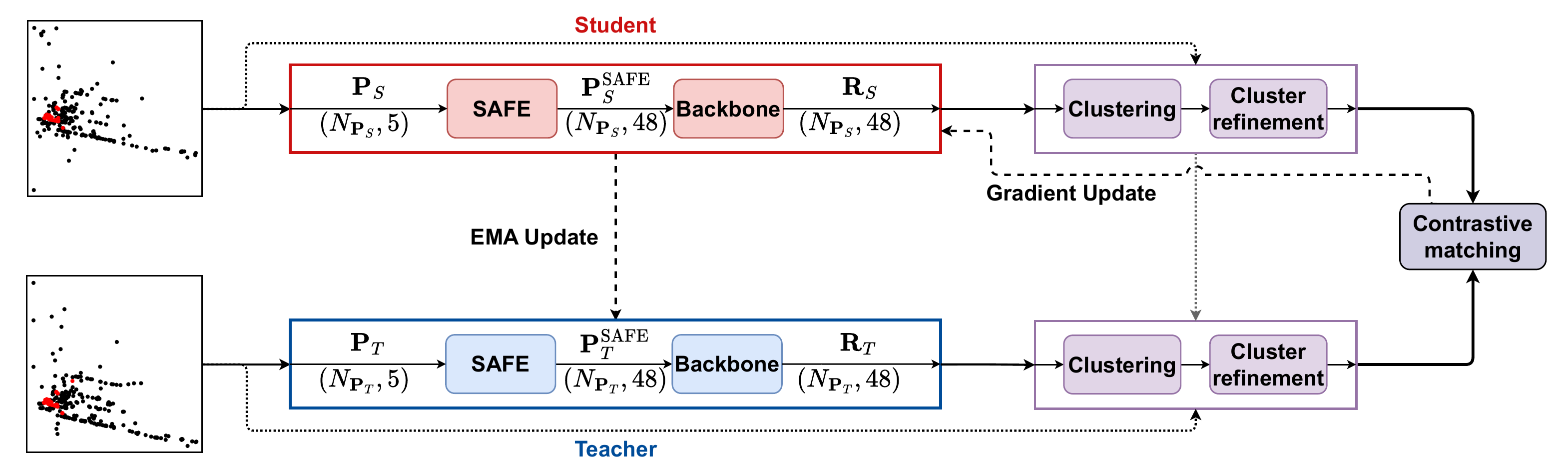}
  \caption{Our self-supervised framework uses a student-teacher architecture employing a modified version of the radar instance transformer architecture~\cite{zeller2024tor} together with our proposed contrastive cluster-based L2-loss with cluster refinement using the dynamic points removal~\cite{kellner2013itsc}.}
  \label{fig:overall_framework}
\end{figure*}

%%%%%%%%%%%%%%%%%%%%%%%%%%%%%%%%%%%%%%%%%%%%%%%%%%%%%%%%%%%%%%%%%%%%%%%%%%%%%%%%
\section{Related Work}
\label{sec:related}

Self-supervised training can be divided into two categories: full self-supervised training and self-supervised representation learning with fine-tuning. The different methods to process point cloud data are often similar for radar and LiDAR data, which is why we include approaches for both sensor modalities to give a comprehensive overview.
Full self-supervised approaches are often more complex and, at the same time, not easily transferable to other domains. Self-supervision for scene flow estimation solves moving object segmentation~(MOS) as a byproduct and often relies on nearest neighbor correspondences between points in consecutive frames \cite{baur2021iccv, ding2022ral}. These correspondences can be assumed to exist for data obtained by most LiDAR sensors because of their higher resolution. However, their existence can not be guaranteed for most radar sensors due to the lower resolution, leading to a degradation of these methods. To alleviate this problem, Ding~\etal~\cite{ding2022ral} relax the direct nearest neighbor correspondences by matching them in a probabilistic manner. While this might benefit the performance on scene flow estimation, the resulting MOS performance is far below the current state-of-the-art set by supervised methods~\cite{ding2022ral,zeller2024tor}.

LiDAR-based representation learning often uses the established student-teacher framework for contrastive learning~\cite{caron2021iccv,tarvainen2017nips,haliassos2022cvpr}. Wang~\etal~\cite{wang2024cvpr} train their network to produce semantic-aware representations by extending the basic framework with learnable prototypes and an additional prediction layer behind the student, making the proposed framework asymmetric. Huang~\etal~\cite{huang2021iccv} do not use any learnable prototypes or asymmetry but incorporate a temporal sampler to generate temporally related input pairs for spatio-temporal representation learning. Both methods show a performance increase for the resulting fine-tuned models and demonstrate an increased label efficiency for LiDAR-based semantic segmentation.

Radar-based representation learning adapts contrastive learning, but instead of intra-modal supervision, these methods often use knowledge distillation from the camera domain~\cite{wang2023distillation,hao2024bootstrapping}. Hao~\etal~\cite{hao2024bootstrapping} combine both supervision modalities, using a student-teacher architecture for the intra-modal supervision where both networks directly share their weights and a vision encoder for cross-modal supervision from camera data. They report an increased performance of the resulting fine-tuned model over training the models from scratch, similar to the methods from the LiDAR domain. However, these approaches focus on semantic segmentation and object detection and do not address the task of moving object segmentation. 

Approaches to moving object segmentation of point clouds primarily focus on supervised learning and can roughly be split into four groups, including projection-based, voxel-based, point-based, and transformer-based methods.

Projection-based methods use point cloud projections into range-view or bird's-eye-view images to use image processing networks to segment the points into moving and non-moving~\cite{cheng2024mf,kim2022ral,sun2022iros}. These methods are prominent in LiDAR-based MOS, leading to state-of-the-art results on the SemanticKITTI moving object segmentation benchmark~\cite{chen2021ral}. Due to the lower resolution of the radar, however, these methods often degrade in performance for radar-specific tasks~\cite{zeller2023ral}.

Voxelization-based methods~\cite{wang2023insmos,sun2020pointmoseg,mersch2022ral} often build upon sparse 4\,D convolutions, first proposed by Choy~\etal~\cite{choy2019cvpr} to incorporate temporal information directly in the spatial domain of LiDAR point clouds. Unfortunately, the voxelization needed to obtain the sparse tensors on which the convolutions operate introduces discretization artifacts and leads to information loss, which may limit their performance on sparse radar point clouds~\cite{zeller2023icra}.

A way to avoid these discretization artifacts and information loss is to work with point-based methods such as PointNet++~\cite{qi2017nips}, FLOT~\cite{puy2020flot}, and FlowNet++~\cite{wang2020wacv}. The approaches focus on scene flow estimation, requiring two consecutive scans to estimate the scene flow and resulting motion segmentation, increasing the runtime and memory consumption over single-scan methods.

The last group of approaches builds upon transformer architectures~\cite{pang2024radarmoseve,zeller2023icra,zeller2024tor}. Each of these approaches uses a different network architecture, but every method has central attention-based aspects to exploit the affinity of neighboring points. These approaches focus on radar-based moving object segmentation or moving instance segmentation in a fully supervised manner. To improve the performance and solve the task of moving object segmentation, we propose an optimized self-supervised learning approach.

%%%%%%%%%%%%%%%%%%%%%%%%%%%%%%%%%%%%%%%%%%%%%%%%%%%%%%%%%%%%%%%%%%%%%%%%%%%%%%%%
\section{Our Approach}
\label{sec:main}

We follow a two-step approach to self-supervised pretraining using contrastive learning and subsequent supervised fine-tuning of the model using limited amounts of annotated data. The overall goal of our contrastive pretraining is to train the network to encode the representations of points belonging to the same class close to each other while pushing the representations of the two classes as far apart as possible. This is done by computing the loss between positive and negative pairs of moving and non-moving points by exploiting clustering and an algorithmic approach to motion segmentation.

\subsection{Overall Framework}
For our overall pretraining framework shown in \figref{fig:overall_framework}, we choose a student-teacher framework consisting of two networks sharing the same architecture, but not their parameters. The student's parameters~$\theta$ are updated using normal gradient updates via backpropagation, while the teacher's parameters~$\theta'$ are an exponential moving average (EMA) of the student's parameters, defined as 
\begin{equation*}
    \theta'_{t+1} = (1-\alpha)\,\theta'_t + \alpha\,\theta_{t+1}
\end{equation*} 
with a decay factor of~$\alpha$. This approach has been shown to stabilize the student during training and facilitate convergence~\cite{caron2021iccv}. Each network gets a point cloud~$\mathbf{P}\in\mathbb{R}^{N\times D}$ as input and produces representation vectors~$\mathbf{R}\in\mathbb{R}^{N\times N_{out}}$ where each~$\mathbf{r}_i \in\mathbf{R}$ is the corresponding representation of the point~$\mathbf{p}_i\in\mathbf{P}$ with~$i\in1\ldots N$. The point cloud $\mathbf{P}$ consists of the point coordinates and point-wise features, including the ego-motion compensated Doppler velocity and the radar cross section, resulting in $D=5$ input channels per point. The produced representations $\mathbf{r}_i \in\mathbf{R}$ each consist of $N_{out}=48$ dimensions.
Based on the input point cloud~$\mathbf{P}_S$, the student network produces the corresponding representations~$\mathbf{R}_S$, and based on the input point cloud~$\mathbf{P}_T$, the teacher network produces the corresponding representations~$\mathbf{R}_T$. The two point clouds~$\mathbf{P}_S$ and $\mathbf{P}_T$ form a sample pair where~$\mathbf{P}_S\neq\mathbf{P}_T$. For a positive sample pair, the two point clouds are consecutively recorded frames from the same sequence, while for a negative sample pair, the point clouds are frames from different sequences.

\subsection{Network Architecture}
As the underlying architecture, we chose the radar instance transformer~(RIT) proposed by Zeller~\etal~\cite{zeller2024tor}, which achieves state-of-the-art performance in moving object segmentation. It was originally designed for radar-based moving instance segmentation, which combines moving object segmentation with a separation of the moving points into individual instances. The RIT uses previous scans to enrich the current radar point cloud in the sequential attentive feature encoding~(SAFE) module to reduce computational complexity. The enriched point cloud is processed by the backbone, which consists of transformer blocks. To customize the architecture for the sub-task of moving object segmentation, we modify the original network by removing the instance transformer head and only keeping the SAFE module and the backbone for the pretraining step. While the original RIT takes previous scans into account, we only use the current scan to solve the moving object segmentation on a single-scan basis. For the subsequent fine-tuning step, we append a small multilayer perceptron to generate the motion segmentation masks from the representations produced by the backbone. For more details on the RIT architecture, we refer to the original paper~\cite{zeller2024tor}.

\subsection{Motion-Aware Contrastive Loss Function}

\begin{figure*}[tb]
    \centering
    \includegraphics[width=0.98\linewidth]{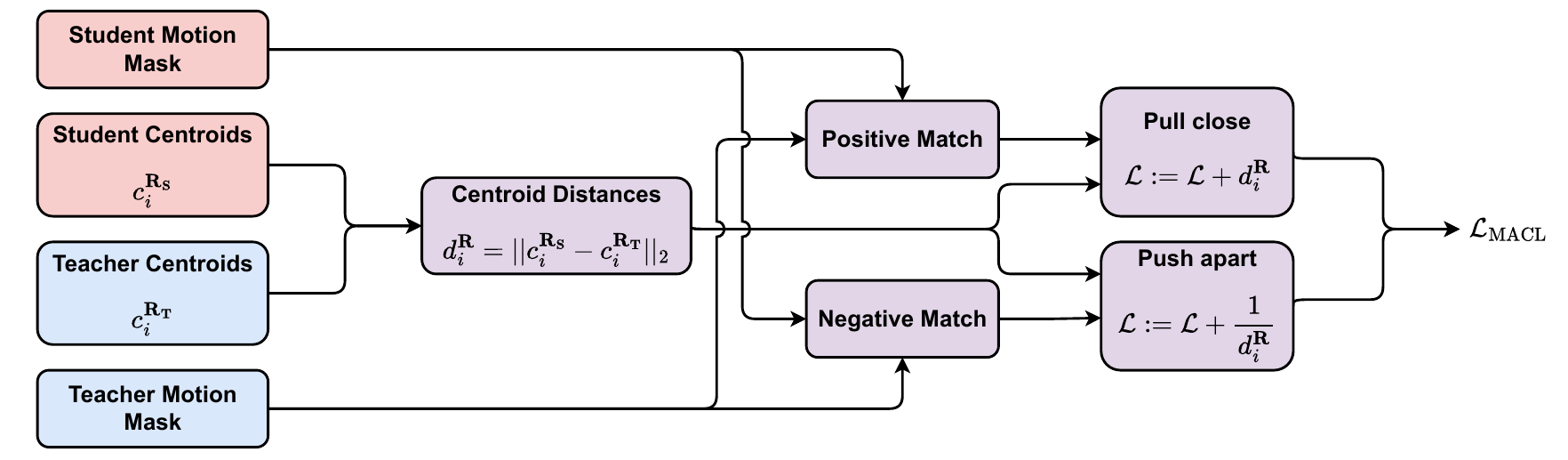} 
    \caption{Illustration of the final process of loss calculation given the representation space centroids and dynamic points removal motion segmentation masks for the student and teacher inputs}
    \label{fig:loss_calculation}
\end{figure*}

The overall goal of our pretraining is to learn feature representations of the data that make the subsequent classification of points into moving and non-moving points easier.
To achieve that goal, our contrastive loss function is supposed to make the network represent the feature encodings of points belonging class close to each other while pushing the different classes apart. In our loss function, we compare the representations produced by the student network with the representations produced by the teacher network based on their respective input. At first, we group the points of both input point clouds into individual clusters and refine those clusters based on pseudo-labels generated using their Doppler velocity~\cite{kellner2013itsc}. After this refinement, every cluster contains only moving or static points, as determined by the pseudo-labels. Then, we match the clusters between the student's and the teacher's input and take the generated pseudo-labels to decide whether the match is positive or negative. For a positive match, both clusters have to belong to the same class estimated by the pseudo-labels, i.e., both clusters contain only moving or only static points. The match is negative if one cluster is moving while the other is static.
To integrate this information into the loss function, we sum up the Euclidean distances between positive matches in the representation space, leading to a minimization of their distance as the loss function is minimized. For negative matches, we sum up the inverse of their distance in the representation space, leading to a maximization of their distance as the loss function is minimized.
This process of matching and contrasting the different clusters is illustrated in \figref{fig:loss_calculation} and embodies the contrastive nature of the loss function and our overall approach.

For the clustering of the student input, we use a variant of HDBSCAN, namely HDBSCAN($\hat{\epsilon}$)~\cite{malzer2020mfi}, which effectively acts as a hybrid of standard DBSCAN and HDBSCAN. It can discover clusters of widely varying densities, but at the same time, avoid the high number of micro-clusters HDBSCAN tends to produce. In this clustering, we only use the three spatial dimensions and the Doppler velocity, but not the radar cross-section, as it can differ by a high margin even for points belonging to the same object. This yields the student cluster label mapping~${L_S: \mathbf{P}_S \to (\mathbb{N} \cup \{-1\})}$, assigning each point $\mathbf{p}_S\in\mathbf{P}_S$ in the student point cloud a cluster label~$L_S(\mathbf{p}_S)$. The label~$-1$ corresponds to points classified as noise by the clustering algorithm. Those points are not used for any of the following steps.
To make the matching of the clusters easier, we force the number of clusters in the teacher and student point clouds to be the same. Consequently, we derive the cluster mapping for the teacher~$L_T$ from the one calculated for the student. We assign to each point in the teacher input point cloud the label of the student cluster whose centroid is closest with respect to the three spatial dimensions calculated based on the Euclidean distance. 

We group the points into clusters and take the distance of their respective centroids for the loss calculation, rather than taking the spatial distance between individual 3\,D points, to reduce the computational load because of fewer comparisons. At the same time, this abstracts from properties of individual points and forces the network to focus on general trends in the data instead.

The resulting clusters are refined using a motion segmentation mask estimated by the dynamic points removal filtering~(DPR) proposed by Kellner \etal~\cite{kellner2013itsc}. DPR uses RANSAC to create a velocity profile describing the relationship between the azimuth of a point and its Doppler velocity. As the majority of points are considered stationary, every point deviating from this profile by more than a specific threshold is considered moving. We utilize this motion segmentation mask to refine the previously calculated clusters, resulting in the refined label mappings~$L'_S$ and~$L'_T$. During the refinement step, we split the clusters that contain both moving and static points, as estimated by the DPR, by moving all static points into a new cluster. In this way, each cluster, initially containing moving and static points, is divided into two disjoint clusters, where one keeps the original cluster label and contains all of the moving points, and the other has a new cluster label and contains all of the static points from the original cluster. This step is done independently for the student and the teacher, but it has to be noted that if the $i$-th cluster is split for both student and teacher, the corresponding new clusters containing the static points get the same new label. As a direct consequence of the initial label derivation for the teacher and the mentioned property of common labels for static clusters after refinement, the set of common cluster labels for student and teacher, denoted by~$L'_S \cap L'_T$, is not empty and can be used to match the clusters for the final loss calculation. After the refinement, we use the resulting cluster labels to calculate the centroids of the clusters in the representation space by transferring a point's cluster label to its corresponding representation. Based on these labels, we calculate the representation space centroid for each cluster and use the centroids for the final loss calculation. The final motion-aware contrastive loss~$\mathcal{L}_\mathrm{MACL}$ is defined as 
\begin{equation*}\label{eq:ssl_loss}
    \mathcal{L}_\mathrm{MACL} = \frac{1}{|L'_S \cap L'_T|}\sum_{i \in L'_S \cap L'_T} \mathcal{L}_i,
\end{equation*}
where the individual cluster losses are as follows
\begin{equation*}\label{eq:cluster_loss}
    \mathcal{L}_i = 
        \begin{cases}
            d_i^\mathbf{R} & \quad \text{if } l_i^{\mathbf{R}_S} \text{ and } l_i^{\mathbf{R}_T} \text{ both moving or both static} \\
            \frac{1}{d_i^\mathbf{R}} & \quad \text{else}
        \end{cases}
\end{equation*}

Here, the clusters $l_i^{\mathbf{R}_S}$ and $l_i^{\mathbf{R}_T}$ are the representations that are assigned the cluster label $i$, and $d_i^\mathbf{R}$ is the distance of their representation space centroids.

\subsection{Implementation Details}
We implemented our method in PyTorch~\cite{paszke2019neurips} and leverage the HDBSCAN($\hat{\epsilon}$) implementation provided by McInnes~\etal~\cite{mcinnes2017joss}. 

We pretrain the network for 100 epochs and use the SGDW~\cite{loshchilov2018iclr} optimizer with a momentum~$\mu$ of 0.9 and a weight decay~$\lambda$ of 0.01. The initial learning rate is set to 0.001 and is scheduled using a multi-step scheduler, dropping the learning rate by a factor of 10 at epochs 60 and 80. For the clustering, we use a minimal cluster size of 2 points and a cluster selection epsilon of 0.1. The RANSAC threshold used for the DPR filtering is set to 0.5.

For fine-tuning the network, we follow a similar training regime as the original RIT. Since we modified the architecture by replacing the instance transformer head with a small MLP, we also have to adapt the original loss function by only keeping the focal Tversky loss~\cite{abraham2019isbi} for the semantic output and removing the binary cross-entropy, which was used to train the similarity matrices involved in the instance association. We fine-tune the network utilizing the AdamW~\cite{loshchilov2018iclr} optimizer with a weight decay of 0.01 and an initial learning rate of 0.001 that is dropped by a factor of 10 at epochs 60 and 80 by a multi-step scheduler. To reduce overfitting, we use different augmentations like random rotation, shifting, scaling, and flipping. Furthermore, we do not freeze the pretrained parameters of the radar instance transformer.

All of the pretraining and fine-tuning is done on one Nvidia A100 GPU with a batch size of 128. The only exception is fine-tuning the model using only 1\% of the View-of-Delft training data, since that corresponds to only 51 training samples. In this case, we use a batch size of 8.

%%%%%%%%%%%%%%%%%%%%%%%%%%%%%%%%%%%%%%%%%%%%%%%%%%%%%%%%%%%%%%%%%%%%%%%%%%%%%%%%
\section{Experimental Evaluation}
\label{sec:exp}

The main focus of this work is an approach to self-supervised contrastive representation learning supporting radar-based moving object segmentation.
We present our experiments to show the capabilities of our method. The results of our experiments also support our key claims, which are:
(i) Our approach is able to improve the label efficiency of the underlying network architecture when compared to training the model from scratch;
(ii) Using only 1\% of annotations from the View-of-Delft dataset, our approach outperforms the self-supervised baseline of RaFlow~\cite{ding2022ral} by over 25 absolute percentage points;
(iii) The clustering, as well as the cluster refinement using the dynamic points removal filtering, are essential to successfully group and differentiate the classes of moving and non-moving points in the representation space and thereby improve the performance.

\subsection{Experimental Setup}

We perform our experiments on the View-of-Delft dataset~\cite{palffy2022ral}, which offers approximately 8600 radar point clouds. We transform the existing 3D bounding box labels provided by the dataset into point-wise annotations to efficiently evaluate the segmentation results. To illustrate the generalization capabilities of our method, we also perform experiments on the RadarScenes dataset~\cite{schumann2021fusion}, which is the only publicly available large-scale radar dataset providing point-wise annotations for moving objects in different scenarios. Following Zeller~\etal~\cite{zeller2023ral}, we use the originally defined training set of 130 sequences while splitting the original test set into six sequences for validation (sequences: 6, 42, 58, 85, 99, 122) and the remaining 22 sequences for the evaluation.
In total, these 158 sequences amount to approximately 211000 radar point clouds.

When fine-tuning the models pretrained by our approach, we take different fractions of the dataset's training data to better evaluate the network's label efficiency. For the View-of-Delft dataset, we use fractions of 1\%, 10\%, and 100\%, and for the RadarScenes dataset, because of its larger size, we use fractions of 0.1\%, 1\%, 10\%, and 100\% of the available training data. For validation and testing, we use the aforementioned subsets.

We use the established Intersection-over-Union (IoU), defined as $\text{IoU} = \frac{TP}{TP + FP + FN}$ with the number of true positive ($TP$), false positive ($FP$), and false negative ($FN$) predictions, as the metric to quantify the performance of a method. The IoU is calculated for each class separately, but our main focus is the IoU of the moving class. For completeness, we also provide the static IoU as well as the mean IoU over both classes.

%%%%%%%%%%%%%%%%%%%%%%%%
\subsection{Moving Object Segmentation Performance}

\begin{table}[tb]
    \centering
    \begin{tabular}{lrrrr}
        \toprule
        Model & Data Fraction & Mean IoU & Static IoU & Moving IoU \\
        \midrule
        RIT* & 1\% & 69.2 & 97.4 & 41.1 \\
        \shaderow Ours & 1\% & \textbf{71.8} & \textbf{97.6} & \textbf{46.0} \\
        \midrule
        RIT* & 10\% & 75.7 & 97.9 & 53.4 \\
        \shaderow Ours & 10\% & \textbf{77.4} & \textbf{98.1} & \textbf{56.6} \\
        \midrule
        RaFlow & 100\% & 55.4 & 89.9 & 21.0 \\
        RVT & 100\% & 75.2 & 97.5 & 52.9 \\
        RIT* & 100\% & 79.2 & 98.3 & 60.2 \\
        \shaderow Ours & 100\% & \textbf{81.0} & \textbf{98.5} & \textbf{63.5} \\
        \bottomrule
    \end{tabular}
    \caption{Moving object segmentation results on the View of Delft validation set after finetuning the pretrained backbone or training from scratch with different fractions of the available annotated training data. RIT* is optimized for moving object segmentation.}
    \label{tab:resultsVoD}
\end{table}

The first experiment evaluates the performance of our approach on the View-of-Delft dataset, and its outcomes support the claims that (i) our approach improves the label efficiency of the underlying network architecture and (ii) outperforms the self-supervised approach of RaFlow by 25\% moving IoU using only 1\% of the available training data for finetuning.

As summarized in \tabref{tab:resultsVoD}, our approach outperforms the base architecture of the modified RIT when it is trained from scratch without our self-supervised pretraining. The modified RIT, denoted as RIT*, consists of the originally proposed SAFE module together with the original RIT backbone, but instead of the instance transformer head used for moving instance segmentation, we append a small multilayer perceptron to determine the moving object segmentation mask based on the representations produced by the backbone. 

Throughout the different data fractions, the self-supervised pretraining leads to a performance improvement of more than 3\% moving IoU. \tabref{tab:resultsVoD} also shows that our approach achieves a moving IoU that is 25\% higher than the performance achieved by the self-supervised baseline of RaFlow~\cite{ding2022ral} even when we use only 1\% of the available annotated training data. This amounts to just 51 annotated radar scans, which is a scale manageable to obtain for most entities working on radar-based sensing. Considering that RaFlow was developed to address the task of scene flow estimation, this observation not only motivates approaches dedicated to the task of moving object segmentation but also suggests that, as soon as at least a minimal amount of annotated data is available, it is worth considering methods employing self-supervised pretraining with supervised fine-tuning. To give further context for comparison to other approaches, we show the performance achieved by the radar velocity transformer (RVT)~\cite{zeller2023icra}, which is the recently best-performing transformer-based approach to radar-based single-scan moving object segmentation. The modified RIT architecture trained from scratch already outperforms the RVT by more than 7\% moving IoU, but our approach achieves a moving IoU of 63.5\% when using 100\% of the available training data for the fine-tuning, which is an additional 3\% more than the RIT trained from scratch.

\begin{table}[tb]
    \centering
    \begin{tabular}{lrrrr}
        \toprule
        Model & Data Fraction & Mean IoU & Static IoU & Moving IoU \\
        \midrule
        RIT* & 0.1\% & 77.8 & \textbf{98.2} & 57.3 \\
        \shaderow Ours & 0.1\% & \textbf{78.2} & \textbf{98.2} & \textbf{58.1} \\
        \midrule
        RIT* & 1\% & 88.2 & \textbf{99.1} & 77.2 \\
        \shaderow Ours & 1\% & \textbf{88.9} & \textbf{99.1} & \textbf{78.7} \\
        \midrule
        RIT* & 10\% & 90.4 & \textbf{99.3} & 81.6 \\
        \shaderow Ours & 10\% & \textbf{90.5} & \textbf{99.3} & \textbf{81.8} \\
        \midrule
        RVT & 100\% & 90.3 & 99.3 & 81.3 \\
        RIT* & 100\% & \textbf{91.7} & \textbf{99.4} & 84.0\\
        \shaderow Ours & 100\% & \textbf{91.7} & \textbf{99.4} & \textbf{84.1} \\
        \bottomrule
    \end{tabular}
    \caption{Moving object segmentation results on the RadarScenes test set after finetuning the pretrained backbone or training from scratch with different fractions of the available annotated training data. RIT* is optimized for moving object segmentation.}
    \label{tab:resultsRS}
\end{table}

The second experiment evaluates the performance of our approach on the RadarScenes dataset and illustrates the generalization capabilities of our method.
Unfortunately, the RadarScenes dataset does not provide image data which is necessary for the dataset preparation required for the RaFlow approach. For this reason, we do not include RaFlow in the evaluation on the RadarScenes dataset.

\tabref{tab:resultsRS} shows the results of evaluating our approach as well as the modified RIT and the RVT trained from scratch on the above-defined test partition of the RadarScenes dataset. It can be seen that our approach consistently outperforms the other methods, although the margin of performance gain compared to the modified RIT trained from scratch is smaller than for the experiments performed on the View-of-Delft dataset. This can be explained by the larger size of the RadarScenes dataset. The larger amount of data available for the training of the supervised methods counterbalances the improvements achieved by our self-supervised pretraining to some extent.

%%%%%%%%%%%%%%%%%%%%%%%%
\subsection{Ablation Study on Loss Components}

\begin{table}[t]
    \centering
    \begin{tabular}{lp{1cm}rrr}
        \toprule
        Backbone & Data Fraction & Mean IoU & Static IoU & Moving IoU \\
        \midrule
        w/o DPR & \multicolumn{1}{r}{1\%} & 71.7 & \textbf{97.6} & 45.8 \\
        w/o Clustering & \multicolumn{1}{r}{1\%} & 67.8 & 97.0 & 38.5 \\
        Ours & \multicolumn{1}{r}{1\%} & \textbf{71.8} & \textbf{97.6} & \textbf{46.0} \\
        \midrule
        w/o DPR & \multicolumn{1}{r}{10\%} & 77.0 & \textbf{98.1} & 55.8 \\
        w/o Clustering & \multicolumn{1}{r}{10\%} & 76.1 & 98.0 & 54.2 \\
        Ours & \multicolumn{1}{r}{10\%} & \textbf{77.4} & \textbf{98.1} & \textbf{56.6} \\
        \midrule
        w/o DPR & \multicolumn{1}{r}{100\%} & 79.3 & 98.3 & 60.4 \\
        w/o Clustering & \multicolumn{1}{r}{100\%} & 79.3 & 98.2 & 60.3 \\
        Ours & \multicolumn{1}{r}{100\%} & \textbf{81.0} & \textbf{98.5} & \textbf{63.5} \\
        \bottomrule
    \end{tabular}
    \caption{Influence of the components of the cluster-based L2-loss with DPR refinement on the actual MOS results trained and evaluated on View-of-Delft}
    \label{tab:ssl-loss-ablation-mos}
\end{table}

Finally, we perform an ablation study on the components of our proposed cluster-based L2-loss with DPR refinement to support our third claim and show that each component of the loss function contributes to the performance improvement in terms of IoU by improving the quality of the pretrained representations. The combined results of the ablation study on the validation set of VoD are shown in \tabref{tab:ssl-loss-ablation-mos}. When removing the DPR refinement from the loss function, shown in \tabref{tab:ssl-loss-ablation-mos} as w/o DPR, we only use the cluster labels determined by the HDBSCAN($\hat{\epsilon}$) algorithm to match the different clusters and calculate their centroids' distances. This effectively removes the motion awareness from the learned representations and only offers spatial information to be incorporated into the representations. When removing the clustering from the loss function, shown in \tabref{tab:ssl-loss-ablation-mos} as w/o Clustering, we effectively consider the two groups of moving and static points as estimated by the DPR to be clusters and cross-match them between the student and the teacher branch of the framework. The moving cluster of the student is positively matched with the moving cluster of the teacher and negatively matched with the static cluster of the teacher. In the same way, we match the static cluster of the student positively with the static cluster of the teacher and negatively with the moving cluster of the teacher. This removes the spatial awareness from the representations and only keeps the motion awareness provided by the DPR.

The results of our ablation study illustrate the effectiveness of our approach in combining spatial and motion information to learn motion-aware representations of 4\,D radar data. While the removal of motion-awareness by removing the DPR seems to be less harmful than removing the spatial information provided by the clustering, both ablations perform noticeably worse than our proposed loss concept.

In summary, our evaluation suggests that our method
provides competitive performance when only a small set of annotated data is available for fine-tuning the network. For a small-scale annotated dataset, we can improve the network's performance by pretraining it with a larger amount of unannotated data recorded with the same sensor setup.

%%%%%%%%%%%%%%%%%%%%%%%%%%%%%%%%%%%%%%%%%%%%%%%%%%%%%%%%%%%%%%%%%%%%%%%%%%%%%%%%
\section{Conclusion}
\label{sec:conclusion}

In this paper, we presented a novel clustering-based contrastive loss function that can be used for self-supervised pretraining to produce motion-aware representations from radar data. Our loss function combines spatial and temporal information to provide context for the network to learn representations that already respect motion information. To achieve that, our loss function uses clustering and an algorithmic approach to motion segmentation to isolate moving and stationary points in the representation space produced by the network and create a form of motion-aware representations. 
We fine-tuned the pretrained network on varying amounts of annotated data to evaluate the resulting label efficiency and compared it to the same architecture trained from scratch, as well as other existing approaches to support the claims made in this paper. The experiments show that the self-supervised pretraining leads to a performance improvement when training on very limited amounts of annotated data, when compared to training the network from scratch. Furthermore, our approach outperforms other completely self-supervised approaches by a large margin, even when using only a few annotated training samples for the fine-tuning.
Overall, our approach shows that self-supervised pretraining can effectively reduce the amount of manual data annotation necessary to achieve comparable performance and can make autonomous systems safer by improving the performance of machine learning methods employed in the system's environment perception.

%%%%%%%%%%%%%%%%%%%%%%%%%%%%%%%%%%%%%%%%%%%%%%%%%%%%%%%%%%%%%%%%%%%%%%%%%%%%%%%%

\bibliographystyle{IEEEtran}

\bibliography{IEEEabrv,glorified,new}

\end{document}